\documentclass[11pt]{article}

\def\aclmode{preprint}

\usepackage[\aclmode]{acl}

\makeatletter
\newif\ifanonymous
\def\@reviewmode{review}
\ifx\aclmode\@reviewmode \anonymoustrue \else \anonymousfalse \fi
\makeatother

\usepackage{times}
\usepackage{latexsym}
\usepackage[T1]{fontenc}
\usepackage[utf8]{inputenc}
\usepackage{microtype}
\usepackage{inconsolata}

\usepackage{graphicx}
\usepackage{booktabs}
\usepackage{float}

\title{One Step at a Time: Trading LLM Autonomy for Process Predictability}

\ifanonymous
  \author{Anonymous}
\else
  \author{Hans Schabert \\
    Amazon Web Services \\
    \texttt{schabert@amazon.com} \\\And
    Christoph Peters \\
    University of the Bundeswehr Munich \\
    \texttt{christoph.peters@unibw.de}}
\fi

\begin{document}

\maketitle

\begin{abstract}
Organizations automating operational processes need more than a correct outcome: they need to predict how a process will run, know which one actually ran, and inspect it step by step. When an agent is the executor that predictability is normally lost---the prescribed procedure goes into the system prompt, and only a final answer comes back. We deliver the procedure step by step over the Model Context Protocol (MCP) instead: a server releases one step at a time, the agent executes it, and each step returns a structured \texttt{step\_output}. This trades autonomy for \emph{predictability}---the agent no longer chooses its own path---and two properties then follow by construction, independent of which executor runs the process. The execution path is prescribed \emph{before} the run, so the process is predictable in advance rather than reconstructed afterwards; and the completed step records form a machine-readable execution log that downstream tooling can audit and optimize step by step.

Evaluating 15,475 trials across 13 SOP-Bench domains and four open-weight executors spanning frontier (Kimi~K2.5), large (DeepSeek~V3.2, GLM 5), and lightweight (Ministral~3~8B) capability, we find step-level delivery makes the executed process predictable and inspectable for \emph{every} executor, and additionally raises accuracy when the executor is small. Across all four, process adherence rises significantly (76--95\% to 95--99\%) and \emph{ungrounded} answers---correct outputs produced without executing the SOP---near-vanish, falling from 2.1--4.5\% to 0.2--0.3\% of trials (per-model reductions of $-$1.9 to $-$4.2pp; all 95\% CIs exclude zero); under prompt-based delivery, 31--49\% of correct answers on \texttt{know\_your\_business} bypass the SOP entirely, even for the frontier executor. Accuracy is where the executor's capability enters: the lightweight executor gains +6.5pp grounded accuracy because supplying the process externally removes a reconstruction burden it cannot carry, while capable ones trade a small raw-accuracy decrement for a predictable, auditable process.
\end{abstract}

\section{Introduction}

On an industrial KYC-style benchmark domain, 31--49\% of an LLM agent's correct answers---including 48\% of a frontier model's (Table~\ref{tab:kyb_ungrounded})---are produced without executing a single one of the prescribed verification tools. Standard accuracy metrics count every one of these as a success. For organizations automating the Standard Operating Procedures (SOPs) that govern operational work---classifying dangerous goods, triaging customer complaints, detecting referral fraud~\cite{detreville2005sop, dumas2018bpm}---this is the deployment barrier in miniature: the answer can be right while the process that was supposed to produce it never ran, and nothing in the final answer reveals the difference. Deployments stall in regulated environments for exactly this reason: organizations cannot predict how an agent will behave, cannot see which steps it actually executed, and cannot attribute failures to actionable root causes~\cite{kaur2022trustworthy}; delegation research identifies trust, control, and explainability among the key factors governing whether humans delegate to agentic systems~\cite{strunk2024delegate}.

The prevailing approach embeds the full SOP in the system prompt and permits the model to execute autonomously~\cite{yao2022react, chase2022langchain}. This introduces a fundamental problem: \emph{the model must reinvent the process from text at every invocation}---parse the procedure, determine which step comes next, select the correct tool and parameters, interpret the result, and decide how to proceed.

On SOP-Bench~\cite{nandi2025sopbench}---an industrial benchmark of 2,000+ tasks across 12 business domains---this reinvention produces two failure modes that undermine trust. First, \emph{behavior is unpredictable}: a lightweight open-weight executor produces on average 20 distinct tool-call sequences per domain (up to 44 on a single procedure; Tables~\ref{tab:path_detail} and~\ref{tab:problem}, which also report the size-robust dominant-sequence share), so no supervisor can anticipate which path the next execution will follow; reasoning is opaque, and failures are undiagnosable short of forensic analysis of the full conversation. Second, and more insidiously, \emph{the model can hallucinate the process itself}: it produces the right answer while behaving as if it executed a procedure it never ran---the \texttt{know\_your\_business} failure that opens this paper, where up to half of a frontier model's correct answers bypass the prescribed verification tools entirely.

\paragraph{Approach.} We propose externalizing the process structure. Rather than embedding the SOP in the system prompt, we deliver it step-by-step via the Model Context Protocol (MCP)~\cite{anthropic2025mcp}. The server controls pacing: it releases one step instruction at a time, the model executes that step, and the server advances only after the model reports completion. The model's autonomy is scoped to the current step.

This constitutes a deliberate trade of autonomy for predictability. The model forfeits the ability to cross-reference future steps or self-correct against the full procedure, and we measure the price of that forfeit: capable executors' raw accuracy declines slightly, and branching domains that require look-ahead lose grounded accuracy outright. In return the system gains three properties addressing the trust prerequisites: \emph{predictable behavior} (the step decomposition fixes the execution path before the run, so what the agent will do is known in advance, and invented paths collapse toward one dominant sequence for capable executors), \emph{verifiable reasoning} (each step declares its inputs and outputs as a structured record, a machine-readable audit trail no prompt-based approach provides), and \emph{diagnosable failures} (that record localizes divergence to a single step, enabling step-level rather than wholesale SOP revision).

\paragraph{Contributions.} We evaluate this approach on 15,475 trials across 13 SOP-Bench domains, four models spanning a capability range (a frontier open-weight executor, Kimi~K2.5; two large open-weight executors, DeepSeek~V3.2 and GLM 5; and a lightweight open-weight executor, Ministral~3~8B), and two conditions (RFC~2119 SOP in system prompt vs.\ step-level MCP delivery); all figures below are detailed in \S\ref{sec:results}. Our contributions are: (1)~We introduce \emph{Grounded TSR}---accuracy conditioned on process adherence---and show that 2.1--4.5\% of trials under prompt delivery (Table~\ref{tab:problem}; up to 31--49\% of correct answers on individual domains) are correct without executing the SOP---standard accuracy metrics systematically overstate process compliance; step-level delivery drives this \emph{ungrounded} rate to 0.2--0.3\% for every model (\S\ref{sec:res_ungrounded}; all 95\% CIs exclude zero; partly structural by design). (2)~We decompose behavioral predictability into \emph{process adherence} and \emph{path consistency} and show step-level delivery improves adherence for every model and consolidates execution toward a dominant path for capable executors. (3)~We find the \emph{accuracy} effect---unlike the auditability gains in (1) and (2), which hold across the lineup---is capability-dependent: the lightweight open-weight executor gains +6.5pp grounded accuracy (Table~\ref{tab:v2_tsr}; its adherence rising from 76\% to 95\%, Table~\ref{tab:v2_adherence}) as externalized process control substitutes for capability, while frontier and large open-weight executors trade a small raw-accuracy decrement ($-$2.5 to $-$4.7pp) for full process visibility. (4)~Because every step declares its inputs and outputs, the per-step record is an optimization substrate as well as an audit trail: it drives a targeted revision cycle that repaired an SOP defect in under a minute, and it surfaced two benchmark-environment defects invisible to prompt-based evaluation.

\section{Background and Related Work}

\subsection{SOP-Bench}

SOP-Bench~\cite{nandi2025sopbench} provides 2,000+ tasks from human expert-authored SOPs across 12 business domains (healthcare, logistics, finance, content moderation, trust \& safety, media), each with executable tool interfaces, precomputed mock API responses, and human-validated ground truth; the public release we evaluate contains 14 domains (2,163 tasks). The benchmark reports Execution Completion Rate (ECR), Task Success Rate (TSR), and Conditional TSR (C-TSR).

Their Function-Calling and ReAct baselines show that newer models do not guarantee better performance (Claude~4~Opus: 72.4\% vs.\ Claude~4.5~Sonnet: 63.3\% task success rate on ReAct), and that no model--agent combination dominates (best performances range 57--100\% by domain).

These findings motivate our central hypothesis: the bottleneck is not procedural complexity but \emph{instructional specificity}---and the delivery mechanism determines how faithfully a model adheres to even well-specified instructions.

\subsection{RFC~2119 Formalization}

RFC~2119~\cite{bradner1997rfc2119} defines keywords (MUST, MUST NOT, SHALL, SHOULD, MAY) for indicating requirement levels in technical specifications. Originally designed for internet protocol standards, these keywords establish a precise vocabulary for distinguishing absolute requirements from recommendations.

Prior work compiles ambiguous SOPs into plans and code templates~\cite{giroh2025syntact}; we keep the SOP as prose and formalize only its \emph{normative} force with RFC~2119 keywords. The hypothesis is that explicit normative keywords reduce ambiguity (``You MUST call \texttt{calculate\_sds\_label\_score}\ldots'' vs.\ ``calculate the safety data sheet score''). Our bridge study (Table~\ref{tab:bridge}) confirms this: RFC~2119 formalization lifts DeepSeek~V3.2's raw TSR from 61\% to 83\% (+21.8pp) on 12 domains, though not all of this lift is grounded in actual process execution (+17.8pp grounded; Table~\ref{tab:bridge}).

However, keyword insertion alone is insufficient: keyword-only conversion produced negligible improvement in early experiments. The effective conversion adds (1)~exact tool-to-step mappings with parameter names, (2)~decision thresholds, and (3)~explicit edge-case handling---consistent with AgentIF~\cite{qi2025agentif}, where even the best model follows fewer than 30\% of agentic instructions perfectly, and a substantial share of conditional-constraint failures stem from incorrect condition checking rather than action execution.

\subsection{SOP-MCP: Step-Level Delivery via MCP}

MCP~\cite{anthropic2025mcp} provides a standardized interface for connecting LLMs to external tools and data sources. We use its tool-based delivery mechanism to implement step-by-step SOP execution: a \texttt{run\_sop} tool releases one instruction at a time, the model executes the step, and the server advances only after the model reports a \texttt{step\_output}. The server controls pacing---the model cannot see ahead.

With the full SOP in the system prompt, the model has \emph{SOP-level autonomy}: it can cross-reference steps, reorder operations, and self-correct. With step-level delivery, autonomy is scoped to the \emph{current step}: the model executes each instruction but cannot compensate for errors in the SOP itself. This trade-off---less autonomy for more predictability---is the central design choice we evaluate.

\subsection{Trust and Delegation}

Delegation requires predictable behavior and verifiable compliance. Prompt-based SOP execution fails this test: with dozens of distinct execution paths for the same procedure (up to 44 in our data; Table~\ref{tab:path_detail}), no supervisor can predict the next execution and no auditor can verify compliance. MCP delivery addresses this directly: the server enforces a defined path, and step-level documentation provides the evidence trail auditors require.

This connects to a broader literature on autonomy--control trade-offs in human organizations~\cite{detreville2005sop}: strict checklists improve consistency but reduce workers' ability to catch errors in the checklist itself. The same tension applies to LLM agents; we quantify both sides in Section~\ref{sec:results}.

\subsection{Decomposition: Advisory vs.\ Enforced}

Our approach---externalizing process control so the model executes one step at a time---relates to a family of decomposition strategies in which planning and execution both remain inside the model's own forward pass, whether by planning then solving, routing sub-problems to handlers, building up from simpler sub-questions, or using a controller LLM to sequence tools~\cite{wang2023planandsolve, khot2023decomposed, zhou2023leasttomost, shen2023hugginggpt}. Closest to our setting, FlowBench~\cite{xiao2024flowbench} benchmarks workflow-guided planning with workflow knowledge supplied in several formats, and ToolPlanner~\cite{wu2024toolplanner} trains for path planning with process feedback; step-level process signals have also been used to \emph{train} agents~\cite{xiong2024watcheverystep}, whereas we use them to revise the \emph{procedure}.

All of these share the intuition that decomposition improves reliability, and in all of them the model retains control over sequencing---the decomposition is \emph{advisory}. In our MCP-based approach the server controls sequencing---the decomposition is \emph{enforced}---and it is this enforced variant whose consistency and auditability we measure in Section~\ref{sec:results}.

\section{Methodology} \label{sec:methodology}

\subsection{Experimental Design}

We compare two SOP delivery conditions across four models and 13 industrial domains from SOP-Bench.

\paragraph{Conditions.}
\begin{enumerate}
    \item \textbf{rfc2119\_sop}: The SOP converted to RFC~2119 normative format with explicit tool mappings, decision thresholds, and MUST/SHOULD/MAY keywords, delivered as a system prompt. The model has full SOP-level autonomy.
    \item \textbf{sop\_mcp}: The same RFC~2119 SOP decomposed into discrete steps and delivered one at a time via the \texttt{run\_sop} tool. The server controls pacing---the model executes each step, reports a \texttt{step\_output}, and receives the next instruction. The model's autonomy is scoped to the current step.
\end{enumerate}

Both conditions use identical SOP content. The only difference is the delivery mechanism: bulk system prompt vs.\ step-by-step MCP.

\paragraph{Models.} Four open-weight models span a capability range to test capability dependence (Table~\ref{tab:models}), all accessed via Amazon Bedrock's Converse API with a single code path (July--August~2026). List pricing and full per-model economics appear in Appendix~\ref{sec:appendix_economics}.

\begin{table}[ht]
\centering
\small
\setlength{\tabcolsep}{4pt}
\begin{tabular}{lllr}
\toprule
\textbf{Executor} & \textbf{Version} & \textbf{Tier} & \textbf{max\_tok} \\
\midrule
Ministral & 3 8B & light & 4,096 \\
DeepSeek & V3.2 & large & 8,192 \\
GLM & 5 & large & 8,192 \\
Kimi & K2.5 & frontier & 8,192 \\
\bottomrule
\end{tabular}
\caption{Executors: family name (used throughout), version, capability tier, and output-token cap. Bedrock model IDs: \texttt{mistral.ministral-3-8b-instruct}, \texttt{deepseek.v3.2}, \texttt{zai.glm-5}, \texttt{moonshotai.kimi-k2.5}.}
\label{tab:models}
\end{table}

All models are invoked with provider-default sampling (we set neither temperature nor top\_p; notably not temperature=0) and the max\_tokens caps of Table~\ref{tab:models}; observed generation sits well below both caps (Table~\ref{tab:token_usage}). Each task is run once per model$\times$condition (one trial per task; the MCP condition uses a multi-turn step loop within one session), so bootstrap resamples draw over distinct tasks and involve no within-task clustering.

\paragraph{Domains.} SOP-Bench provides 14 domains. We exclude one (\texttt{warehouse\_package\_inspection}): a pre-existing bug in SOP-Bench's source code causes its \texttt{validateBarcode} tool to compute barcode match via PO-number parity rather than ground-truth values, making faithful SOP execution systematically incorrect in 41\% of cases (Section~\ref{sec:warehouse_traceability}). Of the remaining 13, all are run; \texttt{video\_annotation} is excluded from adherence metrics (wide branching defeats the fixed threshold; Limitations). The 13 evaluated domains span content moderation, healthcare, finance, retail, media, logistics, and trust \& safety, totaling 15,475 trials; 12 form the primary adherence pool.

\subsection{Bridge Study: Connecting SOP-Bench to Our Evaluation}

Our study uses RFC~2119 SOPs and a different agent framework (Strands + Bedrock Converse) than SOP-Bench's evaluation (LangChain + invoke\_model, Claude~3.5~v2). To validate comparability, we ran DeepSeek~V3.2---an open-weight model whose DeepSeek-R1 sibling SOP-Bench reports---across all 14 domains and four conditions (plain SOP, plain + tool appendix, RFC~2119, SOP-MCP; 8,618 trials).

Table~\ref{tab:bridge} shows both raw TSR and \emph{Grounded TSR}---accuracy restricted to trials where the model actually followed the SOP (pa\_coverage $\geq 0.5$, 12 domains excluding \texttt{video\_annotation}). The gap between raw and grounded reveals how much accuracy comes from process shortcuts.

\begin{table}[ht]
\centering
\small
\begin{tabular}{lrrrr}
\toprule
\textbf{Condition} & \textbf{TSR} & \textbf{Grounded} & \textbf{$\Delta$} & \textbf{Ungr\%} \\
\midrule
plain\_sop & 60.8\% & 60.6\% & +0.2pp & 0.3\% \\
plain\_enriched & 61.5\% & 61.2\% & +0.2pp & 0.4\% \\
rfc2119\_sop & 82.6\% & 78.4\% & +4.2pp & 5.1\% \\
sop\_mcp & 78.0\% & 77.8\% & +0.1pp & 0.2\% \\
\bottomrule
\end{tabular}
\caption{Bridge study (DeepSeek~V3.2, 12 domains, domain-mean). Grounded TSR counts only correct answers from adherent trials. $\Delta$ = accuracy from shortcuts. Ungr\% = fraction of correct answers that are ungrounded.}
\label{tab:bridge}
\end{table}

Three findings emerge: (1)~our plain\_sop baseline (61\%) falls inside SOP-Bench's reported per-domain range for its own agent baselines (57--100\%), indicating our harness reproduces comparable behaviour rather than introducing an artifact; (2)~RFC~2119 formalization lifts raw TSR +21.8pp but only +17.8pp grounded---the formalized SOP lets the model infer answers without calling tools: ungrounded answers rise from 0.3\% to 5.1\% of correct answers; (3)~appending tool information to informal SOPs gives negligible lift (+0.7pp raw). Formalization helps, but raw TSR overstates it.

\subsection{SOP-MCP Protocol}

The \texttt{run\_sop} tool delivers one step at a time: each call returns the step instruction, and the model must fully execute it before advancing. The system prompt names the SOP, prohibits improvisation, and requires calling \texttt{run\_sop} first; this suffices across all four models.

\subsection{SOP Quality Assurance}

Our RFC~2119 SOPs required iterative correction---an automated linter and step-trace inspection identified tool-name mismatches, schema/implementation conflicts, and incorrect branching logic across four domains during pilot runs. All corrections occurred before the primary evaluation, and every reported trial for a revised domain comes from a post-correction re-run of its full task set; pilot tasks were not a disjoint development split (Limitations). The debugging signal came exclusively from MCP step traces (no equivalent pass for the prompt condition); this asymmetry is acknowledged as a limitation but also demonstrates step-level traceability's diagnostic value (Section~\ref{sec:warehouse_traceability}).

\subsection{Evaluation}

We evaluate on three dimensions, all computed from OpenTelemetry tool-call traces.

\paragraph{Grounded Task Success Rate.} For each task, the agent's output is parsed from structured tags (\texttt{<final\_decision>}, JSON) and compared against ground truth using fuzzy matching (case-insensitive, prefix-stripping, containment checks). We report \emph{Grounded TSR}: correct answers restricted to trials where the model adhered to the SOP (pa\_coverage $\geq 0.5$), divided by total trials; raw TSR follows by identity (raw = grounded + ungrounded). Effect signs are stable for thresholds 0.3--0.7 (Table~\ref{tab:sensitivity}). Conditioning success on evidence of actual execution is under active development in concurrent work~\cite{flynt2026groundeval, soni2026toolfailbench, huang2026workflowfidelity}; related benchmarks instead verify final environment state~\cite{trivedi2024appworld} or reliability across repeated trials~\cite{yao2025taubench}. We operationalize it for \emph{SOP adherence} and quantify the raw-versus-grounded gap across a capability range. Appendix~\ref{sec:appendix_pa} gives the full pa\_coverage definition.

\paragraph{Process Adherence.} The fraction of trials where the model called at least half the \emph{expected} tools (pa\_coverage $\geq 0.5$). This pragmatic threshold avoids marking correct runs on branching SOPs as non-adherent. The expected set is defined per task: for linear SOPs it is the SOP's prescribed domain tools; for branching SOPs whose multi-field output identifies the branch taken (\texttt{email\_intent}) it is that branch's necessary tools; for branching SOPs with single-field outputs (\texttt{customer\_service}, \texttt{video\_classification}) it is estimated empirically as the tools used by the majority of correct prompt-condition executions of that task. This per-case denominator avoids penalizing legitimate early exits and alternative valid paths. Coverage is name-based and order-insensitive: it verifies \emph{which} tools were called, not their order, parameters, or whether outputs were consumed (Limitations). Trials below the threshold indicate the model bypassed the SOP; we report the \emph{ungrounded} rate---correct answers from non-adherent trials---alongside. (The bridge study, re-run with DeepSeek~V3.2 through the same pipeline, shows the same pattern: ungrounded answers concentrate under rfc2119 delivery and vanish under MCP; Table~\ref{tab:bridge}.)

\paragraph{Path Consistency.} We introduce \emph{dominant tool-sequence percentage} as a behavioral predictability metric, computed only among adherent trials. For each (domain, model, condition) group, we extract the ordered sequence of tool names, count the frequency of each unique sequence, and report the percentage following the most common (dominant) sequence. Higher values indicate more reproducible behavior.

\section{Results} \label{sec:results}

All accuracy results report \emph{Grounded TSR}---correct answers from trials where the model adhered to the SOP (pa\_coverage $\geq 0.5$), divided by total trials---measuring only accuracy backed by an auditable process trace. Results pool trials across 12 domains (\texttt{video\_annotation} excluded: branching makes the threshold unreliable; Limitations); deltas carry trial-level bootstrap CIs (10,000 resamples, 95\%; $n \approx 1{,}880$ per model$\times$condition). GLM 5's grounded delta is aggregation-sensitive (trial-pooled $-$2.9pp vs.\ domain-mean +0.8pp; domain sizes 30--274); all other effects are robust to the aggregation choice.

\subsection{Failure Modes Under Prompt-Based Delivery}

Table~\ref{tab:problem} (Appendix~\ref{sec:appendix_problem}) quantifies the two failure modes under prompt-based delivery that motivate this work. First, \emph{hallucinated process execution}: every model produces correct answers without executing the prescribed tools---2.1--4.5\% of all trials, and on \texttt{know\_your\_business} \emph{31--49\% of each model's correct answers}, including 48\% of the frontier executor's (Appendix~\ref{sec:appendix_kyb}). Standard accuracy metrics count these as successes. Second, \emph{process reinvention}: given the identical SOP, the lightweight executor averages 20 distinct tool-call sequences per domain, with barely half of adherent trials on its most common path. Under these conditions no auditor can certify compliance and no supervisor can predict the next execution.

\subsection{Effect on Ungrounded Answers} \label{sec:res_ungrounded}

Step-level delivery drives the ungrounded rates of Table~\ref{tab:problem} to $\approx$0.3\% for every model: Ministral $-$4.2pp [$-$5.2, $-$3.2], DeepSeek $-$2.9pp [$-$3.8, $-$2.1], Kimi $-$2.1pp [$-$2.9, $-$1.4], GLM~5 $-$1.9pp [$-$2.6, $-$1.2] (all CIs exclude zero); on \texttt{know\_your\_business} the ungrounded share of correct answers collapses from 31--49\% to 0--8\% (Table~\ref{tab:kyb_ungrounded}). This near-elimination is partly structural---the server withholds the next step until the current one executes---and it is measured against coverage-based adherence (Appendix~\ref{sec:appendix_pa}); under a stricter completeness criterion it is robust for the lightweight executor but not identifiable for the capable ones (Limitations). Two findings survive regardless: the problem it removes is large under prompt-based delivery (up to 49\% of correct answers on \texttt{know\_your\_business}), and all four models comply in practice rather than answering from priors.

\subsection{Process Adherence and Path Consistency}

Table~\ref{tab:v2_adherence} reports both predictability dimensions. \emph{Adherence} (trial-pooled) rises significantly for every model, most for the lightweight one (+18.8pp). This holds for any threshold $t \in [0.3, 0.7]$, but requiring \emph{every} expected tool ($t = 1.0$) reverses it for GLM and Kimi (87$\to$84\%, 89$\to$83\%). Among adherent trials, \emph{path consistency}---the fraction following the single most common tool-call sequence, computed per domain and macro-averaged---rises for every model, but under a domain-level bootstrap the gain is significant only for DeepSeek~V3.2 (+13.7pp) and GLM~5 (+7.0pp); Kimi and Ministral gain about +4.5pp with intervals spanning zero. Distinct-path counts shrink for all capable executors (path-count ratios MCP/RFC of 0.48--0.85$\times$). For linear SOPs this approaches determinism: on \texttt{dangerous\_goods}, GLM's paths collapse from 6 to 1 (100\% dominant).

\begin{table}[ht]
\centering
\small
\setlength{\tabcolsep}{3.5pt}
\begin{tabular}{lrrrrrr}
\toprule
& \multicolumn{3}{c}{\textbf{Adherence}} & \multicolumn{3}{c}{\textbf{Consistency}} \\
\cmidrule(lr){2-4} \cmidrule(lr){5-7}
\textbf{Executor} & \textbf{RFC} & \textbf{MCP} & \textbf{$\Delta$} & \textbf{RFC} & \textbf{MCP} & \textbf{$\Delta$} \\
\midrule
Ministral & 76\% & 95\% & +19\rlap{*} & 53\% & 58\% & +5 \\
DeepSeek & 86\% & 97\% & +11\rlap{*} & 65\% & 78\% & +14\rlap{*} \\
GLM & 92\% & 99\% & +7\rlap{*} & 70\% & 77\% & +7\rlap{*} \\
Kimi & 95\% & 99\% & +4\rlap{*} & 74\% & 79\% & +4 \\
\bottomrule
\end{tabular}
\caption{Adherence (trial-pooled; fraction of trials executing at least half the expected tools) and path consistency (dominant tool-sequence \% among adherent trials; per-domain macro-average), 12 domains. All adherence deltas are significant. Consistency deltas carry two-stage cluster-bootstrap CIs (resampling domains, then trials within domain; 10,000 resamples): Ministral +4.6 [$-$1.8, +11.3], DeepSeek +13.7 [+5.8, +23.0], GLM 5 +7.0 [+0.1, +17.9], Kimi +4.5 [$-$0.3, +10.0]. * = CI excludes zero.}
\label{tab:v2_adherence}
\end{table}

\paragraph{The lightweight exception.} One predictability dimension does not transfer to the lightweight executor: step-level delivery makes Ministral \emph{follow} the process nearly as often as the capable executors (95\% vs.\ 97--99\% adherence), but its path consistency barely moves (53\%$\to$58\%) and its distinct-path count \emph{rises} under MCP (path ratio 1.47$\times$). Traces show genuine ordering variance concentrated on branching domains rather than execution noise. Executing the process the same way each run remains capability-bound, like the accuracy effect reported next.

\subsection{Effect on Grounded Accuracy}

Table~\ref{tab:v2_tsr} turns to accuracy, the one dimension where the effect depends on the executor.

\begin{table*}[t]
\centering
\small
\begin{tabular}{llrrrr}
\toprule
& & \multicolumn{3}{c}{\textbf{Grounded TSR}} & \textbf{Raw TSR} \\
\cmidrule(lr){3-5} \cmidrule(lr){6-6}
\textbf{Executor} & \textbf{Tier} & \textbf{RFC} & \textbf{MCP} & \textbf{$\Delta$ [95\% CI]} & \textbf{$\Delta$ [95\% CI]} \\
\midrule
Ministral & light & 61.0\% & 67.5\% & \textbf{+6.5} [+3.4, +9.6]\rlap{*} & +2.3 [$-$0.7, +5.4] \\
DeepSeek & large & 78.6\% & 79.0\% & +0.4 [$-$2.2, +3.0] & $-$2.5 [$-$5.1, $-$0.1]\rlap{*} \\
GLM & large & 84.8\% & 82.0\% & $-$2.9 [$-$5.2, $-$0.5]\rlap{*} & $-$4.7 [$-$6.9, $-$2.4]\rlap{*} \\
Kimi & frontier & 85.5\% & 83.8\% & $-$1.7 [$-$4.0, +0.7] & $-$3.8 [$-$6.0, $-$1.5]\rlap{*} \\
\bottomrule
\end{tabular}
\caption{Grounded and raw TSR deltas by executor (trial-pooled, 12 domains; raw = grounded + ungrounded). Bootstrap 95\% CIs, 10,000 trial-level resamples; * = CI excludes zero.}
\label{tab:v2_tsr}
\end{table*}

The lightweight open-weight executor gains substantially in grounded accuracy (Ministral~3~8B, +6.5pp, significant), while the large open-weight and frontier executors show no grounded gain (DeepSeek and Kimi within noise; GLM 5 slightly negative). The raw column makes the composition explicit: Ministral's \emph{raw} TSR rises only +2.3pp (n.s.)---roughly two-thirds of its grounded gain is \emph{reclassification} of shortcut answers into process-backed ones---while capable executors' raw TSR \emph{declines} (GLM 5 $-$4.7pp, Kimi $-$3.8pp, DeepSeek $-$2.5pp, all significant) as shortcuts are removed without replacement. We consider this the point rather than a caveat: the intervention converts unaudited into audited accuracy. Ministral's gain is \emph{adherence-driven} (76\%$\to$95\%, Table~\ref{tab:v2_adherence}; accuracy among adherent trials roughly unchanged, Appendix~\ref{sec:appendix_decomposition}): the model could execute each step but failed to reconstruct the procedure from the prompt. Capable models (adherence 86--95\%) preserve grounded accuracy and gain visibility. Excluding the two domains with empirically derived adherence denominators (see Evaluation) \emph{strengthens} the pattern: Ministral +8.8pp [+5.5, +12.1], DeepSeek +5.0pp [+2.2, +7.9], GLM 5 +0.7pp and Kimi +0.9pp (both n.s.) on the remaining 10 domains.

\paragraph{Per-domain variation.} Gains concentrate where prompt-based adherence was poor (\texttt{traffic\_spoofing\_detection} +39.8pp averaged over models, \texttt{patient\_intake} +13.3pp, \texttt{know\_your\_business} +7.6pp); losses concentrate in branching decision-tree domains where step-scoped autonomy prevents look-ahead (\texttt{video\_classification} $-$21.6pp, \texttt{referral\_abuse\_detection\_v2} $-$14.0pp). Appendices~\ref{sec:appendix_domains} and~\ref{sec:appendix_consistency} give the full breakdown.

\subsection{Cost, Latency, and Errors}

The lightweight executor is cheapest and fastest (\$0.015 per grounded answer, 17s per trial under MCP) but does not match the frontier executor's grounded accuracy; step-level delivery costs 2.1--2.9$\times$ more, and tool errors stay at 1--3\% of calls. Appendix~\ref{sec:appendix_economics} gives per-model tokens, cost, and latency.

\section{Discussion}

\subsection{The Autonomy--Predictability Trade-off} \label{sec:negative_domains}

The central question is not whether step-level delivery uniformly improves accuracy---it does not: it repairs process errors, not judgement, and a step-scoped model cannot look ahead---but whether the trade of autonomy for predictability is worthwhile and for which models. The effect is \emph{capability-dependent}: all four models near-eliminate ungrounded answers; capable executors consolidate execution paths; the lightweight executor additionally gains grounded accuracy.

The interaction with model capability is instructive. The frontier and large open-weight executors reconstruct the process effectively from the prompt (Kimi~K2.5: 85\% Grounded TSR, 95\% adherence, 74\% path consistency; DeepSeek and GLM 5 similar). For them, step-level delivery does not raise grounded accuracy---but it lifts adherence toward 97--99\%, raises path consistency (significantly for DeepSeek and GLM~5, +7 to +14pp), and produces a step-level audit trail. The frontier executor does not \emph{need} externalized process control for accuracy; the organization deploying it needs the visibility.

The lightweight executor is different. Ministral~3~8B at 61\% Grounded TSR under prompt-based delivery does not fail for lack of domain knowledge---it has the same SOP, tools, and data. It fails to reconstruct a multi-step procedure from a long document: adherence is only 76\%, and on some domains it abandons the SOP in most trials. Step-level delivery removes this reconstruction burden---each instruction is concise, specifies one action, names the exact tool---lifting adherence to 95\% and grounded accuracy by +6.5pp. Here externalized process control partially substitutes for capability.

At $\approx$3.5$\times$ lower cost per grounded answer, this makes the lightweight executor a credible engine for process-grounded automation. What remains capability-bound is precisely scoped: at 67.5\% grounded it does \emph{not} match the larger executors (79.0--83.8\%), most of its grounded gain is shortcut answers becoming process-backed (\emph{raw} gain +2.3pp, n.s.), and it follows the process far more often without executing it reproducibly (Section~\ref{sec:results}).

\subsection{SOP Structure and Delivery Fit}

SOPs in practice range from \emph{sequential} checklists, through \emph{hierarchical} step/sub-step structures that still follow one path, to \emph{flowchart} procedures whose next step depends on the current outcome~\cite{dumas2018bpm}.

Our results reveal an interaction between SOP structure and delivery benefit along two axes: path consolidation and grounded accuracy. For \emph{sequential} SOPs (\texttt{dangerous\_goods}, \texttt{order\_fulfillment}, \texttt{content\_flagging}), step-level delivery achieves near-deterministic execution for capable executors: distinct paths collapse toward one and dominant-path consistency reaches 97--100\%. For \emph{branching} SOPs (\texttt{customer\_service}, \texttt{email\_intent}), path consistency stays low (21--37\%) because the variation reflects legitimate branching---different inputs trigger different decision paths---which step-level delivery cannot (and should not) eliminate.

Grounded-accuracy gains, however, track \emph{adherence repair} more than structure: the largest MCP lifts occur where prompt-based adherence was poor, while branching decision-tree domains that require look-ahead can \emph{lose} grounded accuracy under step-scoped autonomy (per-domain figures in \S\ref{sec:results}). The recent SOP-Maze benchmark~\cite{wang2026sopmaze} likewise finds that SOP structure shapes failure modes.

This suggests a natural scope: step-level delivery is most effective for sequential and hierarchical SOPs---the majority of operational procedures in regulated industries. Extending it to conditional branching, where the server selects the next step from tool outputs, is a clear direction for future work.

\subsection{Traceability Enables Targeted Optimization} \label{sec:warehouse_traceability}

The step-level audit trail is not merely a compliance artifact---because each step declares its inputs and outputs in a fixed shape, it is an interface another program can consume, which is what makes an automated revision cycle possible at all.

\paragraph{Case study: customer\_service.} In pilot runs, the \texttt{customer\_service} domain exhibited a $-$14pp MCP penalty. Step traces revealed the root cause: Steps 9 and 10 contained a decision tree for resolution vs.\ escalation, but the ordering caused the model to match ESCALATED before evaluating RESOLVED. An SOP optimizer agent, given the step-level failure traces, reordered the decision tree (RESOLVED first) within 60 seconds. Following the correction, the pilot MCP penalty was reduced to $-$4pp; the remaining gap traces to a separate conditional-execution issue in Step~9.

This optimization cycle---run trials, identify failing steps from traces, revise those steps, re-run---is feasible only with step-level granularity. Under prompt-based delivery, the same failure manifests as ``wrong answer'' at the trial level, requiring manual conversation analysis across hundreds of trials to locate the root cause.

The same traceability also exposed a benchmark defect in the excluded \texttt{warehouse\_package\_inspection} domain, where higher prompt-based accuracy came from circumventing the tools entirely (Appendix~\ref{sec:appendix_warehouse}).

\subsection{Evidence a Domain Expert Can Check}

\paragraph{Verifiability through step-level evidence.} Every MCP step produces a \texttt{step\_output} documenting what the model did, called, and concluded. This is not obtainable from the full-SOP prompt baseline we evaluate, where the model reasons internally and emits only a final answer; a prompt-based agent could be wrapped to emit comparable checkpoints, but that externalizes the process structure by another route. This is \emph{extrinsic} trust~\cite{jacovi2021trust}: observable evidence of correct behavior, versus impractical \emph{intrinsic} trust in internal reasoning. MCP shifts the trust question from ``Is this model capable?'' to ``Is this process correct?''---answerable by domain experts without ML expertise, and answerable \emph{before} deployment, since the step decomposition fixes the path in advance.

\section{Conclusion}

The barrier to deploying LLM agents is not capability but trust~\cite{strunk2024delegate, kaur2022trustworthy}. Step-level SOP delivery fixes the execution path in advance, raises adherence to 95--99\%, documents every step, and cuts hallucinated process execution to $\approx$0.3\%, at the price of a small raw-accuracy decrement for capable executors.

\section*{Limitations}
\label{sec:limitations}

This study has several limitations.

\begin{enumerate}
    \item \textbf{Bundled mechanisms.} Step-level delivery bundles three mechanisms that this study does not fully disentangle: \emph{information presentation} (the model sees shorter, more specific instructions per turn), \emph{process enforcement} (the model cannot see ahead), and \emph{per-step behavioral meta-instructions} (e.g., ``treat null tool returns as success''). An ablation delivering the SOP in sequential chunks without enforcement---where the model sees one step at a time but can request any step---would isolate these effects; the per-step meta-instructions could similarly be added to the prompt condition to control for their contribution.
    \item \textbf{Mock tools.} SOP-Bench uses mock tools with precomputed results---real-world APIs introduce additional failure modes (timeouts, rate limits, schema changes) not captured here.
    \item \textbf{Sampling temperature.} We do not set \texttt{temperature=0}; while the 15,475-trial sample provides statistical power, individual domain$\times$model cells (as few as 30 trials) have wider confidence intervals.
    \item \textbf{Formalization confound.} The two-condition design (RFC~2119 prompt vs.\ MCP) does not isolate step-level delivery of informal SOPs, which would separate the delivery mechanism from formalization.
    \item \textbf{SOP quality amplification.} MCP amplifies SOP quality in both directions---poorly specified SOPs produce consistent but wrong execution, raising the quality bar for SOP authoring; two benchmark defects we corrected (Section~\ref{sec:methodology}) illustrate both the risk and the diagnostic value of step traces.
    \item \textbf{Domain exclusion.} One SOP-Bench domain was excluded due to a tool--ground-truth mismatch unrelated to our intervention.
    \item \textbf{Adherence threshold.} Our process adherence metric (pa\_coverage $\geq 0.5$) measures coverage against per-case necessary tools where the output identifies the branch, and against empirically derived tool sets for single-output branching domains; \texttt{video\_annotation}'s wide branching still defeats this threshold and is excluded from adherence analysis.
    \item \textbf{Trust is measured indirectly.} We measure the technical infrastructure for trust---adherence, path consistency, grounded accuracy, and step-level evidence---not trust itself. Whether these predictability gains shift actual human delegation decisions remains an open empirical question requiring user studies.
    \item \textbf{Trace-guided SOP debugging.} The SOP debugging cycle used MCP step-level traces, meaning corrections were optimized for defects visible under step-level delivery; an analogous cycle using prompt-based conversation logs might have produced differently structured SOPs. Two further caveats bear on interpretation. First, the pilot runs that surfaced these defects drew on the same SOP-Bench task set used for the reported evaluation, not a disjoint development split, so the corrections may be tuned to these tasks; a held-out split would be the cleaner design. Second, the corrections were frozen before the primary runs in the sense that every reported trial for a revised domain is a post-correction re-run over that domain's full task set --- no reported cell mixes pre- and post-correction trials. The revised SOPs are released with the artifacts so the exact evaluation environment can be reconstructed.
    \item \textbf{Aggregation sensitivity.} Grounded-accuracy deltas can be aggregation-sensitive when domain sizes vary (30--274 tasks): GLM 5's delta is $-$2.9pp trial-pooled but +0.8pp domain-mean; we report trial-pooled figures with matching trial-level CIs and flag this sensitivity explicitly.
    \item \textbf{Name-based coverage, and why a stricter criterion is not identifiable.} Adherence coverage is name-based: it does not verify call order, parameters, or that tool outputs were consumed, so a trial can count as adherent while misusing results. We attempted a stricter \emph{completeness} criterion---did the trial call every tool the task actually required---and found it is not identifiable on this benchmark, because SOP-Bench's ground truth does not enumerate per-task required tools. Every available substitute reference is condition-dependent: deriving the required set from correct prompt-condition trials makes step-level delivery look worse, deriving it from correct step-level trials makes prompt delivery look worse by a similar margin, and the one reference independent of both---mapping ground-truth output fields to the tools that produce them---is available for only two of twelve domains, because the rest have scalar rather than structured ground truth. We therefore report the coverage-based metric, the only definition that generalises across all domains, and note that the ungrounded reduction is robust for the lightweight executor under every definition we tried but metric-dependent for the capable ones. Per-task required-tool annotation is the missing instrument and the clearest next step for this line of work. Separately, the fuzzy output matcher (containment-based) was spot-checked but not human-validated at scale, so verbose answers could in principle produce false-positive matches.
    \item \textbf{Per-domain uncertainty.} Per-domain figures rest on cells of 30--274 tasks without per-domain CIs or multiple-comparison control and should be read as descriptive.
\end{enumerate}

\paragraph{Artifacts.} The RFC~2119 SOPs, step decompositions, system prompts, corrected tool specifications, per-trial OpenTelemetry traces, and analysis code will be released upon publication, so that the evaluation environment (including the two benchmark fixes) can be reproduced exactly. Model runs were launched from per-model configuration files that write into a shared results directory rather than from one combined config, so reproduction proceeds model by model.

\bibliography{references}

\appendix

\section*{Appendix}

\section{Per-Domain MCP Lift} \label{sec:appendix_domains}

Table~\ref{tab:v2_heatmap} shows the grounded MCP lift by domain, averaged across the four models.

\begin{table}[ht]
\centering
\small
\begin{tabular}{lr}
\toprule
\textbf{Domain} & \textbf{Avg MCP $\Delta$} \\
\midrule
traffic\_spoofing\_detection & +39.8 \\
patient\_intake & +13.3 \\
know\_your\_business & +7.6 \\
dangerous\_goods & +3.1 \\
content\_flagging & +2.5 \\
order\_fulfillment & +2.5 \\
\midrule
aircraft\_inspection & $-$2.2 \\
customer\_service & $-$3.4 \\
email\_intent & $-$3.5 \\
referral\_abuse\_detection\_v1 & $-$6.4 \\
\midrule
referral\_abuse\_detection\_v2 & $-$14.0 \\
video\_classification & $-$21.6 \\
\bottomrule
\end{tabular}
\caption{Average grounded MCP lift by domain in percentage points (simple mean of the four per-model Grounded TSR deltas, computed from unrounded per-model values). Gains concentrate where prompt-based adherence was poor (\texttt{traffic\_spoofing\_detection}, \texttt{patient\_intake}); losses concentrate in branching decision-tree domains where step-scoped autonomy prevents look-ahead.}
\label{tab:v2_heatmap}
\end{table}

\section{Grounded TSR and Path Variance: Full Results} \label{sec:appendix_consistency}

Table~\ref{tab:path_detail} reports Grounded TSR for every domain$\times$model$\times$condition cell, with unique execution-path counts in parentheses (computed among adherent trials, pa\_coverage $\geq 0.5$). A cell shows what fraction of trials produced a correct answer \emph{while following the SOP}, and how many distinct tool-call sequences the adherent trials took.

\begin{table*}[t]
\centering
\small
\setlength{\tabcolsep}{4pt}
\begin{tabular}{lrrrrrrrr}
\toprule
& \multicolumn{4}{c}{\textbf{rfc2119\_sop}} & \multicolumn{4}{c}{\textbf{sop\_mcp}} \\
\cmidrule(lr){2-5} \cmidrule(lr){6-9}
\textbf{Domain} & \textbf{M8B} & \textbf{DS} & \textbf{GLM} & \textbf{Kimi} & \textbf{M8B} & \textbf{DS} & \textbf{GLM} & \textbf{Kimi} \\
\midrule
aircraft\_inspection & 79\% \tiny{(25)} & 79\% \tiny{(9)} & 88\% \tiny{(15)} & 86\% \tiny{(8)} & 77\% \tiny{(28)} & 79\% \tiny{(6)} & 80\% \tiny{(4)} & 87\% \tiny{(8)} \\
content\_flagging & 88\% \tiny{(10)} & 89\% \tiny{(5)} & 97\% \tiny{(9)} & 99\% \tiny{(4)} & 89\% \tiny{(18)} & 96\% \tiny{(3)} & 99\% \tiny{(4)} & 99\% \tiny{(4)} \\
customer\_service & 58\% \tiny{(28)} & 85\% \tiny{(22)} & 92\% \tiny{(38)} & 82\% \tiny{(21)} & 70\% \tiny{(60)} & 69\% \tiny{(13)} & 88\% \tiny{(19)} & 76\% \tiny{(13)} \\
dangerous\_goods & 57\% \tiny{(18)} & 95\% \tiny{(12)} & 96\% \tiny{(6)} & 96\% \tiny{(1)} & 78\% \tiny{(15)} & 90\% \tiny{(4)} & 94\% \tiny{(1)} & 93\% \tiny{(4)} \\
email\_intent & 70\% \tiny{(33)} & 86\% \tiny{(7)} & 92\% \tiny{(8)} & 91\% \tiny{(5)} & 73\% \tiny{(16)} & 85\% \tiny{(5)} & 77\% \tiny{(6)} & 91\% \tiny{(5)} \\
know\_your\_business & 31\% \tiny{(23)} & 41\% \tiny{(11)} & 50\% \tiny{(3)} & 41\% \tiny{(7)} & 40\% \tiny{(39)} & 39\% \tiny{(7)} & 56\% \tiny{(4)} & 59\% \tiny{(6)} \\
order\_fulfillment & 80\% \tiny{(2)} & 83\% \tiny{(4)} & 83\% \tiny{(2)} & 87\% \tiny{(1)} & 87\% \tiny{(3)} & 83\% \tiny{(2)} & 87\% \tiny{(2)} & 87\% \tiny{(1)} \\
patient\_intake & 59\% \tiny{(17)} & 88\% \tiny{(5)} & 48\% \tiny{(10)} & 98\% \tiny{(6)} & 70\% \tiny{(15)} & 86\% \tiny{(3)} & 100\% \tiny{(1)} & 91\% \tiny{(3)} \\
referral\_abuse\_detection\_v1 & 95\% \tiny{(7)} & 99\% \tiny{(4)} & 100\% \tiny{(2)} & 99\% \tiny{(1)} & 86\% \tiny{(20)} & 93\% \tiny{(2)} & 94\% \tiny{(1)} & 94\% \tiny{(4)} \\
referral\_abuse\_detection\_v2 & 60\% \tiny{(15)} & 95\% \tiny{(9)} & 94\% \tiny{(8)} & 96\% \tiny{(7)} & 38\% \tiny{(24)} & 83\% \tiny{(4)} & 84\% \tiny{(3)} & 85\% \tiny{(2)} \\
traffic\_spoofing\_detection & 6\% \tiny{(18)} & 9\% \tiny{(5)} & 50\% \tiny{(7)} & 52\% \tiny{(7)} & 64\% \tiny{(29)} & 68\% \tiny{(3)} & 72\% \tiny{(9)} & 72\% \tiny{(3)} \\
video\_classification & 60\% \tiny{(44)} & 81\% \tiny{(14)} & 84\% \tiny{(20)} & 83\% \tiny{(11)} & 44\% \tiny{(85)} & 59\% \tiny{(5)} & 53\% \tiny{(7)} & 65\% \tiny{(14)} \\
\bottomrule
\end{tabular}
\caption{Grounded TSR by domain, model, and condition; parenthesized numbers are unique execution paths among adherent trials. M8B = Ministral~3~8B, DS = DeepSeek~V3.2, Kimi = Kimi~K2.5.}
\label{tab:path_detail}
\end{table*}

\section{Warehouse Package Inspection: When Higher Accuracy Is the Wrong Signal} \label{sec:appendix_warehouse}

The excluded \texttt{warehouse\_package\_inspection} domain illustrates why traceability matters beyond accuracy optimization. In our bridge study (DeepSeek~V3.2, 600 trials), prompt-based delivery achieved 57\% TSR while MCP achieved only 45\%.

Under MCP, the model faithfully followed the SOP: call \texttt{validateBarcode}, branch on \texttt{barcode\_match}, execute downstream tools, and report the tool-determined \texttt{resolution\_status}. Step traces confirmed correct execution in every trial. Yet 55\% were marked incorrect.

The root cause: the benchmark's \texttt{validateBarcode} mock tool determines \texttt{barcode\_match} using PO number parity (\texttt{po\_number \% 2 == 0}), while the ground truth CSV contains independently generated values. These disagree in 62 of 150 test cases (41\%). A perfect tool-following agent would score at most $(150-62)/150 = 58.7\%$; the observed MCP result of 45\% falls below this ceiling due to additional failure sources in downstream tool logic.

The prompt-based model achieved 57\% by circumventing the tools: it read the \texttt{barcode\_match} column directly from task input data and inferred the resolution status without relying on tool outputs. This is pattern matching on input features, not SOP execution. The higher accuracy masked a benchmark defect invisible to prompt-based evaluation.

This illustrates a general principle: \emph{accuracy without traceability can be actively misleading}. In production, the equivalent scenario is an agent that produces plausible outputs by shortcutting the prescribed procedure---passing audits on accuracy while violating process compliance.

\section{Failure Modes Under Prompt-Based Delivery} \label{sec:appendix_problem}

Table~\ref{tab:problem} quantifies the two failure modes that motivate this work (Section~\ref{sec:results}): hallucinated process execution and process reinvention under prompt-based delivery.

\begin{table}[ht]
\centering
\small
\setlength{\tabcolsep}{3.5pt}
\begin{tabular}{lrrrrr}
\toprule
& \multicolumn{3}{c}{\textbf{Ungrounded}} & \multicolumn{2}{c}{\textbf{Reinvention}} \\
\cmidrule(lr){2-4} \cmidrule(lr){5-6}
\textbf{Executor} & \textbf{Ungr}. & \textbf{/Corr} & \textbf{KYB} & \textbf{\#Paths} & \textbf{Cons.} \\
\midrule
Ministral & 4.5\% & 6.8\% & 44\% & 20.0 & 53\% \\
DeepSeek & 3.1\% & 3.8\% & 49\% & 8.9 & 65\% \\
GLM & 2.1\% & 2.4\% & 31\% & 10.7 & 70\% \\
Kimi & 2.3\% & 2.6\% & 48\% & 6.6 & 74\% \\
\bottomrule
\end{tabular}
\caption{Prompt-based delivery (RFC condition, 12 domains). Columns 2--4: correct answers produced without executing the prescribed tools (\% of trials, share of correct, worst-domain share on \texttt{know\_your\_business}). Columns 5--6: mean distinct tool-call sequences per domain among adherent trials, and dominant-sequence consistency.}
\label{tab:problem}
\end{table}

\section{Ungrounded Answers on know\_your\_business} \label{sec:appendix_kyb}

Table~\ref{tab:kyb_ungrounded} backs the per-model figures quoted in the abstract, introduction, and Section~\ref{sec:results}: the fraction of \emph{correct} answers on \texttt{know\_your\_business} that were produced without executing the prescribed verification tools.

\begin{table}[ht]
\centering
\small
\setlength{\tabcolsep}{4pt}
\begin{tabular}{lrrrr}
\toprule
& \multicolumn{2}{c}{\textbf{RFC}} & \multicolumn{2}{c}{\textbf{MCP}} \\
\cmidrule(lr){2-3} \cmidrule(lr){4-5}
\textbf{Executor} & \textbf{Ungr/Corr} & \textbf{(n)} & \textbf{Ungr/Corr} & \textbf{(n)} \\
\midrule
Ministral & 44\% & 22/50 & 8\% & 3/39 \\
DeepSeek & 49\% & 36/73 & 0\% & 0/35 \\
GLM & 31\% & 20/65 & 0\% & 0/50 \\
Kimi & 48\% & 34/71 & 0\% & 0/54 \\
\bottomrule
\end{tabular}
\caption{\texttt{know\_your\_business}: ungrounded correct answers as a share of all correct answers, per model and condition (90 tasks per cell). Under prompt-based delivery, 31--49\% of every model's correct answers bypass the prescribed tools; under step-level delivery this falls to 0--8\%.}
\label{tab:kyb_ungrounded}
\end{table}

\section{Definition of pa\_coverage} \label{sec:appendix_pa}

For a trial, let $C$ be the \emph{set} of distinct tool names the model actually called and $N$ the \emph{set} of expected tool names for that task. Then
\[
\texttt{pa\_coverage} = \frac{|C \cap N|}{|N|},
\]
and the trial counts as adherent when $\texttt{pa\_coverage} \geq 0.5$. Because both sides are sets of names, the metric has the following explicit behaviour, which we state so its limits are unambiguous:

\begin{itemize}
    \item \textbf{Duplicate calls and retries} collapse to one element and neither help nor hurt.
    \item \textbf{Call ordering} is ignored; a trial calling the right tools in the wrong order scores identically to one in the prescribed order.
    \item \textbf{Parameters and tool outputs} are not inspected, so a trial can be adherent while passing wrong arguments or ignoring what a tool returned.
    \item \textbf{Extra tools} outside $N$ are ignored; there is no penalty for calling tools the SOP did not prescribe.
    \item \textbf{Conditional branches} are handled through $N$, which is defined per task rather than per domain: for linear SOPs it is the SOP's prescribed domain tools; where a multi-field output identifies the branch taken (\texttt{email\_intent}) it is that branch's necessary tools; and for single-output branching domains (\texttt{customer\_service}, \texttt{video\_classification}) it is estimated empirically as the tools used by the majority of correct prompt-condition executions of the same task.
    \item \textbf{Empty expected set.} If $N$ is empty the trial is assigned coverage $1.0$ and is therefore vacuously adherent.
\end{itemize}

The metric is thus a coverage proxy for procedural compliance, not a proof of correct execution; Limitations records this, and the two domains whose $N$ is prompt-derived are the ones we exclude in the robustness check reported in Section~\ref{sec:results}. This is a deliberately looser criterion than related benchmarks apply: FlowBench~\cite{xiao2024flowbench} counts an invocation correct only when the tool name and all required parameters match, checked against the ground-truth sequence, and ToolPlanner~\cite{wu2024toolplanner} requires the solution to match the prescribed tool set at the instruction's granularity level. We adopt coverage because SOP-Bench tasks admit multiple valid orderings and legitimate early exits, and because the stricter criterion is not identifiable here (Limitations).

\section{Adherence Threshold Sensitivity} \label{sec:appendix_sensitivity}

Table~\ref{tab:sensitivity} shows Grounded TSR at varying adherence thresholds (pa\_coverage $\geq t$). Effect directions are stable for Ministral (+6.3 to +8.2pp across $t \in \{0.3, 0.5, 0.7\}$), Kimi ($-$1.7 to $-$2.2pp, n.s.), and GLM 5 ($-$2.9 to $-$3.0pp); DeepSeek hovers at zero ($-$0.3 to +0.9pp), crossing sign but never approaching significance.

\begin{table}[ht]
\centering
\small
\begin{tabular}{llrrr}
\toprule
$t$ & \textbf{Executor} & \textbf{rfc2119} & \textbf{sop\_mcp} & \textbf{$\Delta$} \\
\midrule
0.3 & Ministral & 61.3\% & 67.5\% & +6.3pp \\
    & DeepSeek & 79.2\% & 79.0\% & $-$0.3pp \\
    & GLM & 85.0\% & 82.0\% & $-$3.0pp \\
    & Kimi & 86.0\% & 83.8\% & $-$2.2pp \\
\midrule
0.5 & Ministral & 61.0\% & 67.5\% & +6.5pp \\
    & DeepSeek & 78.6\% & 79.0\% & +0.4pp \\
    & GLM & 84.8\% & 82.0\% & $-$2.9pp \\
    & Kimi & 85.5\% & 83.8\% & $-$1.7pp \\
\midrule
0.7 & Ministral & 58.3\% & 66.5\% & +8.2pp \\
    & DeepSeek & 78.1\% & 79.0\% & +0.9pp \\
    & GLM & 84.7\% & 81.6\% & $-$3.0pp \\
    & Kimi & 85.4\% & 83.6\% & $-$1.8pp \\
\bottomrule
\end{tabular}
\caption{Grounded TSR sensitivity to adherence threshold $t$ (trial-pooled, 12 domains). Directions are stable for Ministral, GLM 5, and Kimi; DeepSeek hovers at zero.}
\label{tab:sensitivity}
\end{table}

\section{Grounded TSR Decomposition} \label{sec:appendix_decomposition}

Grounded TSR = adherence rate $\times$ accuracy among adherent trials. Table~\ref{tab:decomposition} decomposes the aggregate results. Acc$|$Adh declines under MCP for all models---a \emph{reclassification} effect: step-level delivery pulls previously non-adherent trials (disproportionately hard cases the model used to abandon or shortcut) into the adherent population. Grounded TSR rises only where the adherence gain outweighs this dilution (Ministral +18.8pp adherence), stays flat where the two roughly cancel (DeepSeek), and dips slightly where adherence was already high (GLM 5, Kimi).

\begin{table}[ht]
\centering
\small
\setlength{\tabcolsep}{5pt}
\begin{tabular}{llrrr}
\toprule
\textbf{Executor} & \textbf{Cond.} & \textbf{Adh\%} & \textbf{Acc$|$Adh} & \textbf{GTSR} \\
\midrule
Ministral & rfc2119 & 75.9\% & 80.4\% & 61.0\% \\
Ministral & sop\_mcp & 94.7\% & 71.2\% & 67.5\% \\
 & $\Delta$ & +18.8pp & $-$9.1pp & +6.5pp \\
\midrule
DeepSeek & rfc2119 & 86.5\% & 90.8\% & 78.6\% \\
DeepSeek & sop\_mcp & 97.2\% & 81.3\% & 79.0\% \\
 & $\Delta$ & +10.7pp & $-$9.6pp & +0.4pp \\
\midrule
GLM & rfc2119 & 92.0\% & 92.2\% & 84.8\% \\
GLM & sop\_mcp & 98.9\% & 82.9\% & 82.0\% \\
 & $\Delta$ & +6.9pp & $-$9.3pp & $-$2.9pp \\
\midrule
Kimi & rfc2119 & 94.5\% & 90.4\% & 85.5\% \\
Kimi & sop\_mcp & 98.7\% & 84.9\% & 83.8\% \\
 & $\Delta$ & +4.2pp & $-$5.6pp & $-$1.7pp \\
\bottomrule
\end{tabular}
\caption{Grounded TSR decomposition (trial-pooled, 12 domains). Adh\% = fraction of trials passing the adherence threshold; Acc$|$Adh = accuracy among adherent trials.}
\label{tab:decomposition}
\end{table}

\newpage
\section{Token, Cost, and Latency Economics} \label{sec:appendix_economics}

Step-level delivery re-delivers the accumulated conversation at each step without prompt caching, multiplying input tokens by 2.1--2.9$\times$ and yielding a cost tax of 2.1--2.9$\times$ and latency tax of 2.2--2.8$\times$. Despite this, the lightweight executor under MCP costs \$0.010/trial at 17s---4.3$\times$ cheaper per trial (equivalently $\approx$3.5$\times$ per \emph{grounded} answer) and 4.8$\times$ faster than Kimi under MCP (\$0.044, 81s). Enabling prompt caching would substantially reduce the tax.

\begin{table}[ht]
\centering
\small
\setlength{\tabcolsep}{4pt}
\begin{tabular}{llrrrr}
\toprule
\textbf{Executor} & \textbf{Cond.} & \textbf{Tok (in/out)} & \textbf{Dur (s)} & \textbf{\$/trial} \\
\midrule
Ministral & rfc2119 & 22.9k / 0.8k & 6.5 & \$0.004 \\
Ministral & sop\_mcp & 66.0k / 1.8k & 16.8 & \$0.010 \\
DeepSeek & rfc2119 & 34.4k / 1.2k & 33.4 & \$0.023 \\
DeepSeek & sop\_mcp & 83.7k / 3.3k & 93.9 & \$0.058 \\
GLM & rfc2119 & 31.7k / 0.8k & 32.3 & \$0.021 \\
GLM & sop\_mcp & 68.9k / 1.7k & 74.5 & \$0.045 \\
Kimi & rfc2119 & 29.8k / 1.0k & 36.6 & \$0.021 \\
Kimi & sop\_mcp & 63.4k / 2.0k & 80.8 & \$0.044 \\
\bottomrule
\end{tabular}
\caption{Per-trial economics (trial-pooled, 12 domains). List prices (\$/M in/out): Ministral~3~8B \$0.15/\$0.15, DeepSeek~V3.2 \$0.62/\$1.85, GLM~5 \$0.60/\$2.20, Kimi~K2.5 \$0.60/\$3.00 (all verified against published list prices for the region used).}
\label{tab:economics}
\end{table}

\begin{table}[ht]
\centering
\small
\setlength{\tabcolsep}{4pt}
\begin{tabular}{llrrrr}
\toprule
\textbf{Executor} & \textbf{Cond.} & \textbf{Med.} & \textbf{p90} & \textbf{p99} & \textbf{Max} \\
\midrule
Ministral & RFC 2119 & 614 & 1,539 & 2,603 & 4,815 \\
 & SOP-MCP & 1,413 & 3,357 & 5,790 & 9,814 \\
DeepSeek & RFC 2119 & 1,043 & 2,111 & 2,805 & 6,037 \\
 & SOP-MCP & 3,148 & 5,066 & 6,030 & 8,144 \\
GLM & RFC 2119 & 756 & 1,462 & 1,828 & 2,137 \\
 & SOP-MCP & 1,586 & 2,791 & 3,753 & 4,391 \\
Kimi & RFC 2119 & 972 & 1,814 & 2,756 & 6,824 \\
 & SOP-MCP & 1,809 & 3,495 & 4,386 & 5,064 \\
\bottomrule
\end{tabular}
\caption{Output-token usage per trial (trial-pooled, 12 domains): total generated tokens summed across all responses of a trial. Because a trial spans multiple model responses, any single response -- the unit the \texttt{max\_tokens} cap applies to -- is strictly smaller than the trial total. Median usage sits at 15--77\% of even the smaller 4,096 cap.}
\label{tab:token_usage}
\end{table}

\end{document}